\documentclass[twocolumn,letterpaper]{IEEEAerospaceCLS}  

\usepackage[]{graphicx,amssymb,amsfonts,amsmath}
\usepackage{enumitem}
\usepackage[ruled,vlined]{algorithm2e}
\newcommand{\ignore}[1]{}  

\newcommand{\myreferences}{references}

\begin{document}
\title{Autonomous Situational Awareness for UAS Swarms}

\author{%
Vincent W. Hill\\ 
Department of Aerospace Engineering\\
University of Alabama\\
Tuscaloosa, AL 35487\\
vwhill@crimson.ua.edu
\and 
Ryan W. Thomas\\ 
Department of Aerospace Engineering\\
University of Alabama\\
Tuscaloosa, AL 35487\\
rthomas8@crimson.ua.edu
\and 
Jordan D. Larson\\
Department of Aerospace Engineering\\
University of Alabama\\
Tuscaloosa, AL 35487\\
jdlarson1@eng.ua.edu
\thanks{\footnotesize 978-1-7281-7436-5/21/$\$31.00$ \copyright2021 IEEE}            
}

\maketitle

\thispagestyle{plain}
\pagestyle{plain}

\maketitle

\thispagestyle{plain}
\pagestyle{plain}

\begin{abstract}
This paper describes a technique for the autonomous mission planning of unmanned aerial system swarms. Given a swarm operating in a known area, a central command system generates measurements from the swarm. If those measurements indicate changes to the mission situation such as target movement, the swarm planning is updated to reflect the new situation and guidance updates are broadcast to the swarm. The primary algorithms featured in this work are A* pathfinding and the Generalized Labeled Multi-Bernoulli multi-target tracking method.
\end{abstract}

\tableofcontents

\section{Introduction}
The massive increase in sensor and computational power over the last decade has greatly accelerated the development of autonomous unmanned aerial systems (UAS). The next step in this evolution is to team groups of UAS together in a swarm to accomplish tasks that are either too complicated or too dangerous for one vehicle to undertake. This adds a layer of complexity to the autonomous UAS guidance, navigation, and control (GNC) problem, since now there is a group of agents that must act in concert without human input. 


In a parallel field to GNC, multi-target tracking  (MTT) problems have been studied vigorously over the last two decades. One significant contribution is the development of MTT filters based on Random Finite Sets (RFS), in particular, work done by Mahler and the Vo brothers in developing the Gaussian Mixture Probability Hypothesis Density (GM-PHD) \cite{Vo2006_TheGaussianMixtureProbabilityHypothesisDensityFilter} and eventually the Generalized Labeled Multi-Bernoulli (GLMB) \cite{Reuter2014_TheLabeledMultiBernoulliFilter} filters. The core concept behind RFS filtering is to regard the collection of target states and measurements received as sets of both unknown values and number of elements (cardinality), the probability densities of which can then be propagated through Bayesian methods for finite set statistics \cite{Mahler2003_MultiTargetBayesFiltering}. This is beneficial when the number of targets is generally unknown and there is a chance of missed detections or spurious measurements not related to targets (clutter), which is generally the case during the operation of UAS swarms. In dynamic, high-risk environments targets, now known as swarm agents, are likely to be incapacitated. Additionally new agents can be introduced, mission objectives, referred to as targets in this work, can move, new obstacles can be discovered, and other features indistinct from agent-generated measurements are common. Hence the RFS formulation is natural to oversee swarm operations. The authors have developed several techniques that extend the RFS multi-target tracking filters to guidance, navigation, and planning \cite{JDL_Linares_ION} \cite{JDL_Thomas_ELQR}. This work will build upon these by incorporating dynamic planning and guidance updating based on target movement as well as a UAS dynamics model specifically in the simulation and using an GLMB filter for the swarm state estimation.

The core contributions of this work are as follows. A human mission initiator defines the mission area size, initial agent and target locations, and obstacles, if any. A mission plan for each agent and the swarm as a whole must be autonomously devised and transmitted to the swarm. This is accomplished by a coupled A*/Hungarian algorithm Simultaneous Target Assignment and Trajectory (STAT) optimization technique. To accomplish the A* trajectory optimization, the mission area is discretized and all possible paths through the discrete nodes between the agent and target are checked until a path with the lowest possible cost is found. After the A* trajectories for each agent-target assignment is determined, the Hungarian algorithm determines the optimal assignments based on distance traveled in each A* trajectory. Next, a set of waypoints from each agent's starting location to its target is mapped from the A* grid to the physical mission area and supplied to each agent's local computer. Each agent must generate its own control law to stabilize itself and accomplish its associated guidance command, recognize when it reaches each waypoint, and adjust its guidance and control to get to the next waypoint.

A central command system, which could be envisioned as a high altitude aircraft with a sensor suite and a powerful processor, must track the swarm as it carries out its mission. If the operation is occurring in a dynamic, high-risk environment, some agents are likely to be incapacitated and targets are likely to move. Central command must recognize when this occurs and adjust the swarm guidance to ensure mission completion. This is accomplished purely through measurements, eliminating the need for two-way communications in this application. 

The following sections describe the technical approach used in this work and present simulation results.

\section{Technical Approach}

\subsection{Aircraft Dynamics and Control}
This work utilizes a linear state space model based on the lateral dynamics of a small fixed-wing UAS for the agent dynamics. Body-frame forward velocity $u$ is assumed to be regulated by a longitudinal control system and is pre-set by the human mission initiator. Agent locations are propagated by numerically integrating the body velocities transformed to the global frame via the heading angle. The agent dynamics model is described in equations (1)-(5).

\begin{equation}
	\dot{\vec{x}} = \underline{A} \vec{x} + \underline{B} \vec{u}
\end{equation}

\begin{equation}
	\vec{x} = \begin{Bmatrix}
		v \\
		p \\
		r \\
		\phi \\
		\psi 
	\end{Bmatrix}
\end{equation}

\begin{equation}
	\vec{u} = \begin{Bmatrix}
		\delta_a \\
		\delta_r 
	\end{Bmatrix}
\end{equation}

\begin{equation}
\underline{A} = \begin{bmatrix}
	-2.382 & 0 & -30.1 & 65.49 & 0 \\
	-0.702 & -16.06 & 0.872 & 0 & 0 \\
	0.817 & -16.65 & -3.54 & 0 & 0 \\
	0 & 1 & 0 & 0 & 0 \\
	0 & 0 & 1 & 0 & 0 
\end{bmatrix}
\end{equation}

\begin{equation}	
\underline{B} = \begin{bmatrix}
	0 & -7.41 \\
	-36.3 & -688 \\
	-0.673 & -68.0 \\
	0 & 0 \\
	0 & 0 
\end{bmatrix}
\end{equation}

The agent dynamics are controlled by a full state feedback linear-quadratic-regulator (LQR) controller with $\underline{Q} = \underline{I}_5$ and $\underline{R} = \underline{I}_2$. The augmented system is given in equations (6)-(7) and is discretized at 100 Hz for simulation.

\begin{equation}
	\dot{\vec{x}} = (\underline{A}-\underline{B}\underline{K}) \vec{x} + \underline{B}\underline{K} \vec{x}_{des}
\end{equation}

\begin{equation}
	\vec{x}_{des} = \begin{Bmatrix}
		0 \\
		0 \\
		0 \\
		0 \\
		\psi_{des} 
	\end{Bmatrix}
\end{equation}

The heading command supplied to $\vec{x}_{des}$ is determined from the angle between the agent's current position and the active waypoint, described in equation (8) where $(x_{ag}, y_{ag})$ and $(x_{ta}, y_{ta})$ are the Cartesian coordinates of the agent and target locations, respectively. When the Euclidean distance between the agent and its current waypoint reaches a certain threshold, in this work 50 meters, the active waypoint updates to the next in the list supplied by central command and mapped to the continuous physical domain.

\begin{equation}
	\psi_{des} = -atan2(y_{ta}-y_{ag}, x_{ta}-x_{ag})
\end{equation}

In the current work, target motion is either zero or prescribed via a 2D double integrator dynamic model discretized at 100 Hz for simulation. $\dot{x}$ and $\dot{y}$ are pre-set in the simulations presented here. 
 
%
%
%
%

\subsection{Swarm Guidance}

Central command must be able to autonomously determine the best mission plan for both pre-set and in-progress agent, target, and obstacle configurations. The technique chosen in this work is a coupled A*/Hungarian algorithm routine. 

A* is a pathfinding / graph traversal algorithm originally develeped in the 1960s \cite{astar} for the development of Shakey the Robot \cite{nilsson}. It achieves better performance than earlier methods such as Djikstra's algorithm by use of a heuristic to guide the search. This heuristic is set as the Euclidean distance between the subject node and the end node, the most common heuristic used. The cost of a node is then calculated as the number of nodes traversed to reach the subject node plus the Euclidean distance from the subject node to the end point.

A* naturally handles obstacles potentially present during swarm operations. If a node is an obstacle, the algorithm will ignore it as a potential waypoint location. This work utilizes both a uniformly distributed random number generator and human-prescribed methods to place obstacles. During real-life swarm operations these obstacles could be envisioned as buildings, known enemy locations, or environmental hazards such as terrain or wildfires. 

After optimal trajectories are generated for each agent/target combination, the optimal target assignment algorithm determines which agent is assigned to which target. This is accomplished by selecting the assignment combination with the lowest total Euclidean distance traveled by all the agents, a method known as the Hungarian algorithm \cite{Kuhn_hungarian} \cite{munkres}. If the number of agents does not equal the number of targets, the algorithm will route additional agents to their closest target or disregard the targets farthest away from agents. 

Once the optimal assignments are selected, the A* grid must be mapped to the physical world. A mesh is generated from the known mission area and A* grid, and each grid waypoint supplied by A* is mapped to the physical world using that mesh. Once the area waypoints have been generated, they are supplied to each agent and stored in local memory for use in the individual UAS' guidance and control. Figures 1 and 2 demonstrate the STAT algorithm results from purely deterministic simulations with one and four agents/targets, respectively, in obstacle-dense environments. This demonstrates that the underlying UAS dynamics and swarm initial mission planning techniques are functional.

\begin{figure}\label{OneColumn}
	\centering
	\includegraphics[width=3.25in]{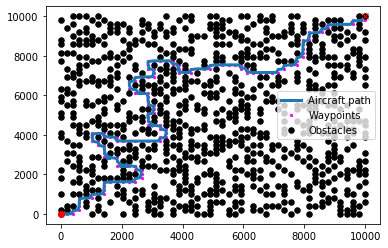}\\
	\caption{\textbf{UAS flight simulation in obstacle-dense environment}}
\end{figure}

\begin{figure}\label{OneColumn}
	\centering
	\includegraphics[width=3.25in]{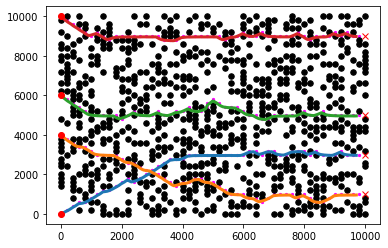}\\
	\caption{\textbf{Four agent/target simulation in obstacle-dense environment}}
\end{figure}

\subsection{Random Finite Set Formulation}
 A Generalized Labeled Multi-Bernoulli (GLMB) random finite set (RFS) multi-target tracking system is used for mission progress monitoring and swarm estimation. A random finite set is a set of random vectors with unknown, time varying cardinality, or number of elements in the set. For the multi-target tracking case the set contains the state vectors for each target and the set's cardinality represents the number of targets tracked. The RFS multi-target tracking literature, developed mostly by the Vo brothers, contains detailed technical descriptions of both the GLMB and predecessor methods [2-6], but a brief development is provided here.
 
 Because the elements of the set, here agent/target states, and the number of elements in the set are unknown, filters that rely on this formulation can also model target birth, spawn, and death. Here birth refers to new targets originating from some modeled probability distribution, here at the beginning of each mission, whereas target spawn refers to new targets created near existing targets. A real-world example of target spawning is an aircraft carrier launching fighters. The probabilistic nature of the RFS also allows the filters to model measurement clutter. A GLMB filter operates on a labeled RFS, which is simply an RFS that has distinct discrete labels appended to each state vector and there is a 1-1 mapping of states and labels. Use of labeled multi-Bernoulli RFSs has been shown to outperform the Probability Hypothesis Density (PHD) \cite{Vo2006_TheGaussianMixtureProbabilityHypothesisDensityFilter} and cardinalized PHD (CPHD) \cite{Vo2006_TheCardinalizedProbabilityHypothesisDensityFilterforLinearGaussianMultiTargetModels} due to its exploitation of both the strengths of the multi-Bernoulli RFS without the weakness of unlabeled MB RFS filters that do not formally track estimates while exhibiting cardinality biases. \cite{Reuter2014_TheLabeledMultiBernoulliFilter}
 
 The Labeled Multi-Bernoulli filter theory was developed by Reuter, B.T. Vo, B.N. Vo, and Dietmayer \cite{Reuter2014_TheLabeledMultiBernoulliFilter} and is summarized below. 
 
 The probability density for a Bernoulli distribution is given by

\begin{equation}
	\pi(X) = \begin{cases}
		1 - r & X = \emptyset \\
		r p(X) & X = \{x\}
	\end{cases}
\end{equation}

where $r$ is the existence probability and $p(\bullet)$ is the spatial distribution. A multi-Bernoulli RFS $X$ is given by the union of the $M$ independent Bernoulli RFSs and is completely defined by the parameter set $\{ (r^{i}, p^{i}) \}_{i=1}^{M}$. The probability density of a multi-Bernoulli RFS can then be described by

\begin{equation}
	\begin{split}
		\pi(\{x_1, \dotsc, x_n\}) &= \prod_{j=1}^{M}(1-r^{(j)}) \times \\
		&\sum_{1 \leq l_1 \neq \dotsi \neq i_n \leq M} \prod_{j=1}^{n}\frac{r^{(i_j)}p^{(i_j)(x_j)}}{1-r^{(i_j)}}\ .
	\end{split}
\end{equation}

In a GLMB each state $x \in$ $\mathcal{X}$ is augmented by a label $l \in$ $\mathcal{L}$ to establish the identity of a single target's trajectory. The labels are ordered pairs of integers where the first index is the time step of birth and the second index is a unique integer to distinguish targets born at the same time step. Finally, the density of a labeled multi-Bernoulli with parameter set $\{ (r^{l}, p^{l}) \}_{l \in L}$ is

\begin{equation}
	\pi({\mathbf{X}}) = \Delta({\mathbf{X}}) w({\mathcal{L}}({\mathbf{X}})) p^{\mathbf{X}}
\end{equation}

where $\Delta({\mathbf{X}}) = 1$ when the cardinality of ${\mathbf{X}}$ equals the cardinality of the labels and the hypothesis weights $w$ are

\begin{align}
	w({\mathcal{L}}({\mathbf{X}})) &= \prod_{i\in {\mathbb{L}}}(1-r^{(i)}) \prod_{\ell \in {\mathbb{L}}} \frac{1_{\mathbb{L}}(\ell) r^{(\ell)}}{1-r^{\ell}} \\
	p^{\mathbf{X}} &= \prod_{\ell \in \mathbb{L}} p^{(\ell)}(x)\ .
\end{align}

Note that the weights and spatial distributions satisfy normalization conditions, summing to one. Here $1_{\mathbb{L}}(\ell)$ is an indicator function, it is equal to one when $\ell \subset \mathbb{L}$ and zero otherwise. The GLMB filter must also solve the measurement association problem to properly propagate the state probability distributions. It mechanizes this by creating a hypothesis for each possible association (including clutter and agent births). Each hypothesis then has a measurement association probability, $w$, associated with it. These hypotheses also track the association history to generate a trajectory, and as such the total number of hypotheses grows with time. For practical use, the computations are kept tractable by pruning hypotheses with low probabilities.

One implementation of this filter is described in detail in \cite{Vo2014_LabeledRandomFiniteSetsandtheBayesMultiTargetTrackingFilter}, and only the main ideas will be summarized here. The filter handles bookkeeping and propagating/updating the hypothesis weights, as well as propagating/updating the underlying single target filters. These underlying single target filters are responsible for propagating and updating each targets state distribution, $p^{(\ell)}$, once the GLMB has determined a measurement association. The underlying filters often take the form of a Kalman filter. This work closely follows the implementation described in \cite{Vo2014_LabeledRandomFiniteSetsandtheBayesMultiTargetTrackingFilter} which uses a Kalman filter to propagate individual states and represents each state probability, $p^{(\ell)}$, as a Gaussian mixture. The underlying filters can then take the form of Bayesian filters, e.g. Kalman filters.

This work utilizes two GLMB filters, one to track agent positions and one to track target positions. So long as the agents and targets do not begin very close to one another each filter will be able to track its desired targets. This is because the birth locations of the agents are outside the birth probability distribution given to the target GLMB and therefore will be treated as clutter by that filter. 
 
\subsection{Algorithm Overview}

The techniques described above were cohered in a single code package to perform numerical simulations. The following provides a basic description of the implementation in this work. Note that specific numeric values should be tailored based on mission requirements.
 
\begin{enumerate}
	\item Initialize agents, targets, and mission region
	\item Generate A* map
	\item For each agent/target combination:
	\begin{enumerate}[leftmargin=5\parindent]
		\item A* search
	\end{enumerate}
	\item For each agent:
	\begin{enumerate}
		\item Optimal target assignment
	\end{enumerate}
	\item Initialize GLMB filter
	\item Main simulation loop:
	\begin{enumerate}
		\item At 100 Hz:
			\begin{enumerate}
			\item For each agent: Propagate dynamics and position
			\item For each target: Propagate dynamics and position
			\item For each agent: If 2D distance to waypoint $<$ 20m, move 
			\item[] on to next
			\item Generate measurement set (agent and target positions 
			\item[] plus clutter)
		\end{enumerate}
		\item At 1 Hz:
			\begin{enumerate}
				\item Execute agent GLMB
				\item Execute target GLMB
			\end{enumerate}
		\item At 0.2 Hz:
		\begin{enumerate}
			\item Associate GLMB outputs with agents and targets
			\item Check if target movement $\geq$ 50m
			\item If true, perform 3(a)-4(a) and 6(a)(iii)
		\end{enumerate}	
	\end{enumerate}
\end{enumerate}

\section{Simulation Results}

The formulation described in Section 2 was implemented in Python in conjunction with two Laboratory for Autonomy, GNC, and Estimation Research (LAGER) public code packages, GNC for Autonomous Swarms Using RFS (GASUR) \cite{gasur} and GNCPy \cite{gncpy}, and used to generate a number of example cases. GASUR was used for its GLMB implementation, while GNCPy was used for its Kalman filter classes.

Several mission simulations, each with four agents and five targets, are presented below. Agents begin each mission on the left side of the map, while targets begin on the right. Obstacles have been placed in the mission area via a uniform random number generator. For each A* node, a uniform random number is generated between 0 and 1. If the number is less than a threshold, 0.25 for this simulation, the node is chosen as an obstacle. Note that targets are allowed to overfly obstacles to present more of a challenge to the swarm guidance computation and that obstacles do not directly generate clutter measurements.  

Figures 3 and 4 demonstrate results of a simulation in which targets move horizontally towards the agent starting locations. Figure 3 plots the states for both the agent and target GLMB RFS filters in black-outlined circles and clutter measurements as transparent gray triangles. Figure 4 shows the true agent and target motion with red o's and x's as initial and final positions, respectively, and obstacles as black dots. These results show that the GLMB filter tracks the agent and target states and that the GLMB outputs can be used to update the swarm guidance during mission execution. 

\begin{figure}\label{OneColumn}
	\centering
	\includegraphics[width=3.25in]{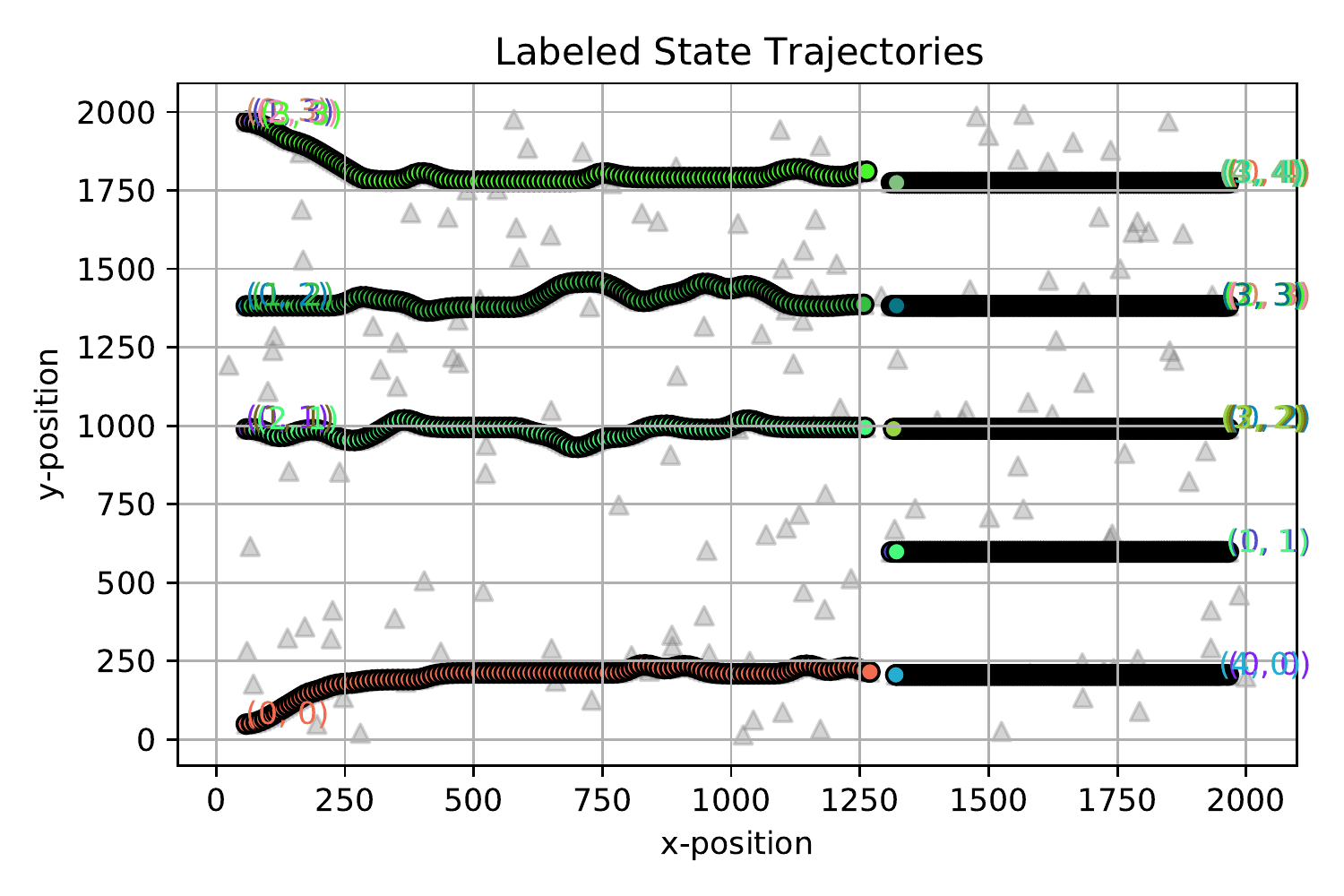}\\
	\caption{\textbf{GLMB output for horizontal target motion}}
\end{figure}

\begin{figure}\label{OneColumn}
	\centering
	\includegraphics[width=3.25in]{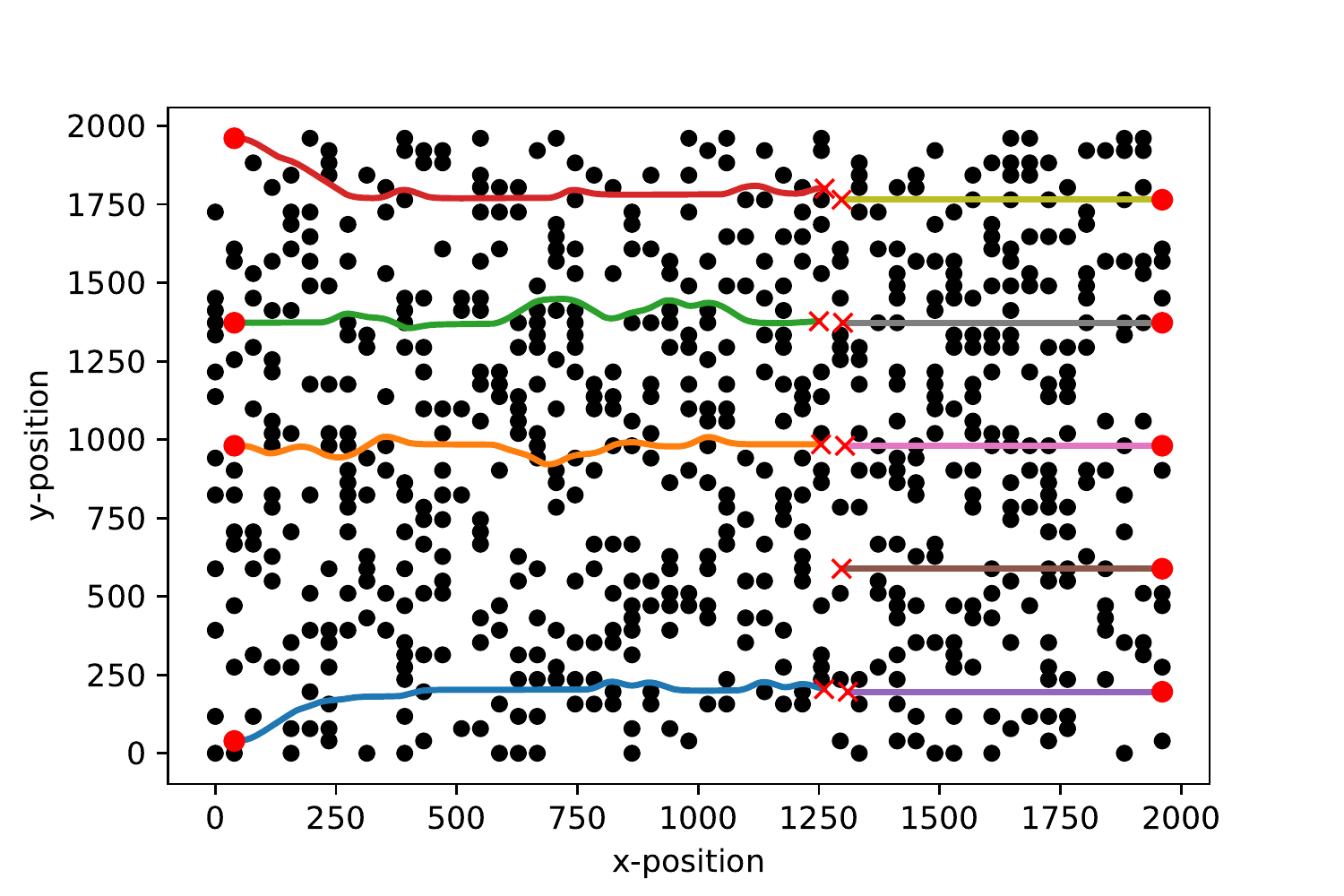}\\
	\caption{\textbf{Ground truth for horizontal target motion}}
\end{figure}

A more interesting and appropriate challenge for this routine is a mission area with high measurement clutter and complex target motion. Figures 5 and 6 present the results of this simulation. Figure 5 exemplifies the robustness of the GLMB to very cluttered measurements. Figure 6 presents the true agent and target trajectories. Waviness in the agent paths is due to the 0.2 Hz A* update rate. This can be ameliorated by increasing that rate, but a compromise must be made between performance and computational requirements. It is clear that this technique can handle high obstacle density with targets moving freely in the mission area. Another similar run was performed with results in Figure 7 and 8. In this simulation target motion and birth locations are randomly assigned via Gaussian distributions. The results in figures 5-8 demonstrate that the technique is able to generate plans for the swarm to complete chaotic missions with high obstacle density, heavily cluttered measurements, and complicated target behavior.

\begin{figure}\label{OneColumn}
	\centering
	\includegraphics[width=3.25in]{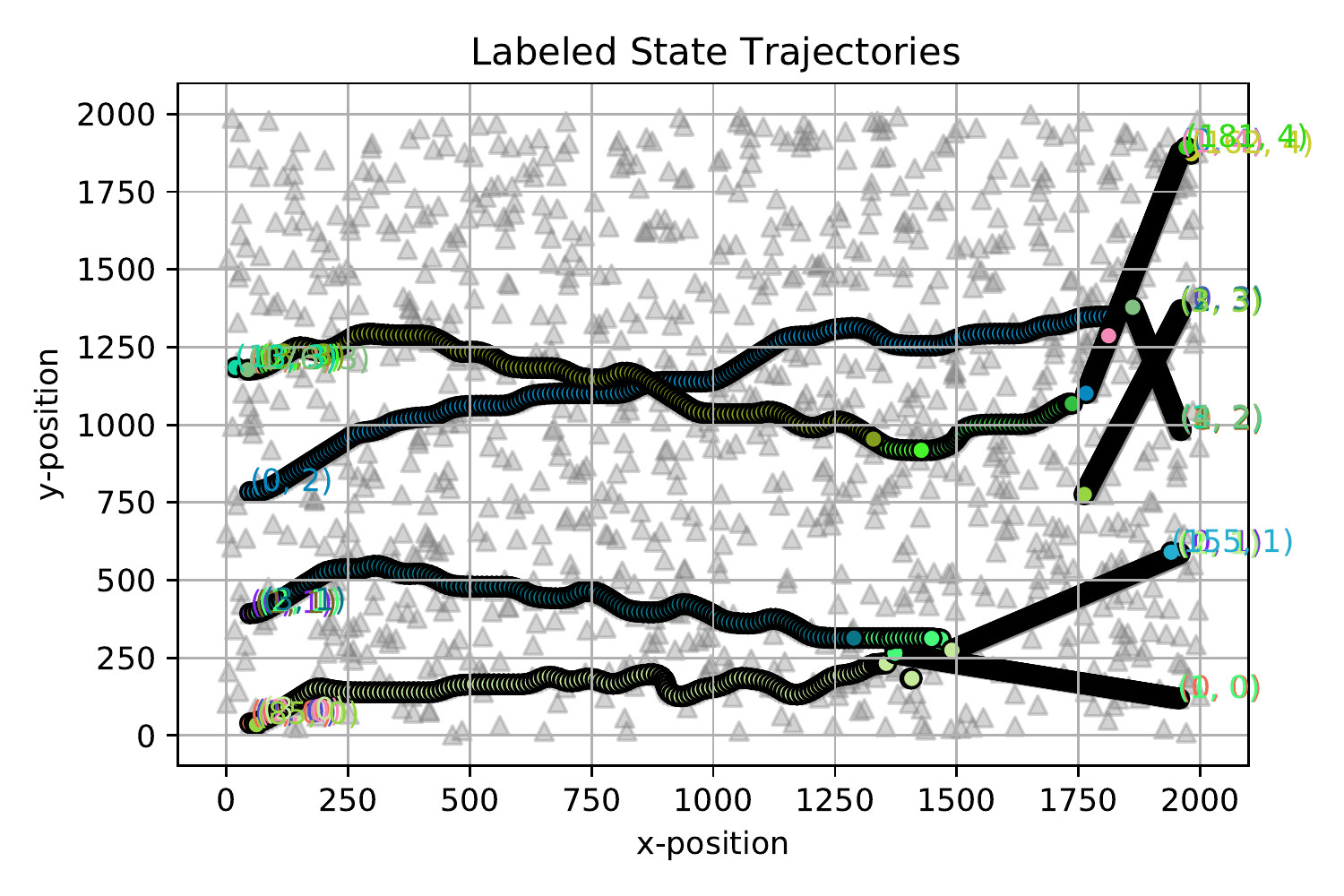}\\
	\caption{\textbf{GLMB output for diagonal target motion}}
\end{figure}

\begin{figure}\label{OneColumn}
	\centering
	\includegraphics[width=3.25in]{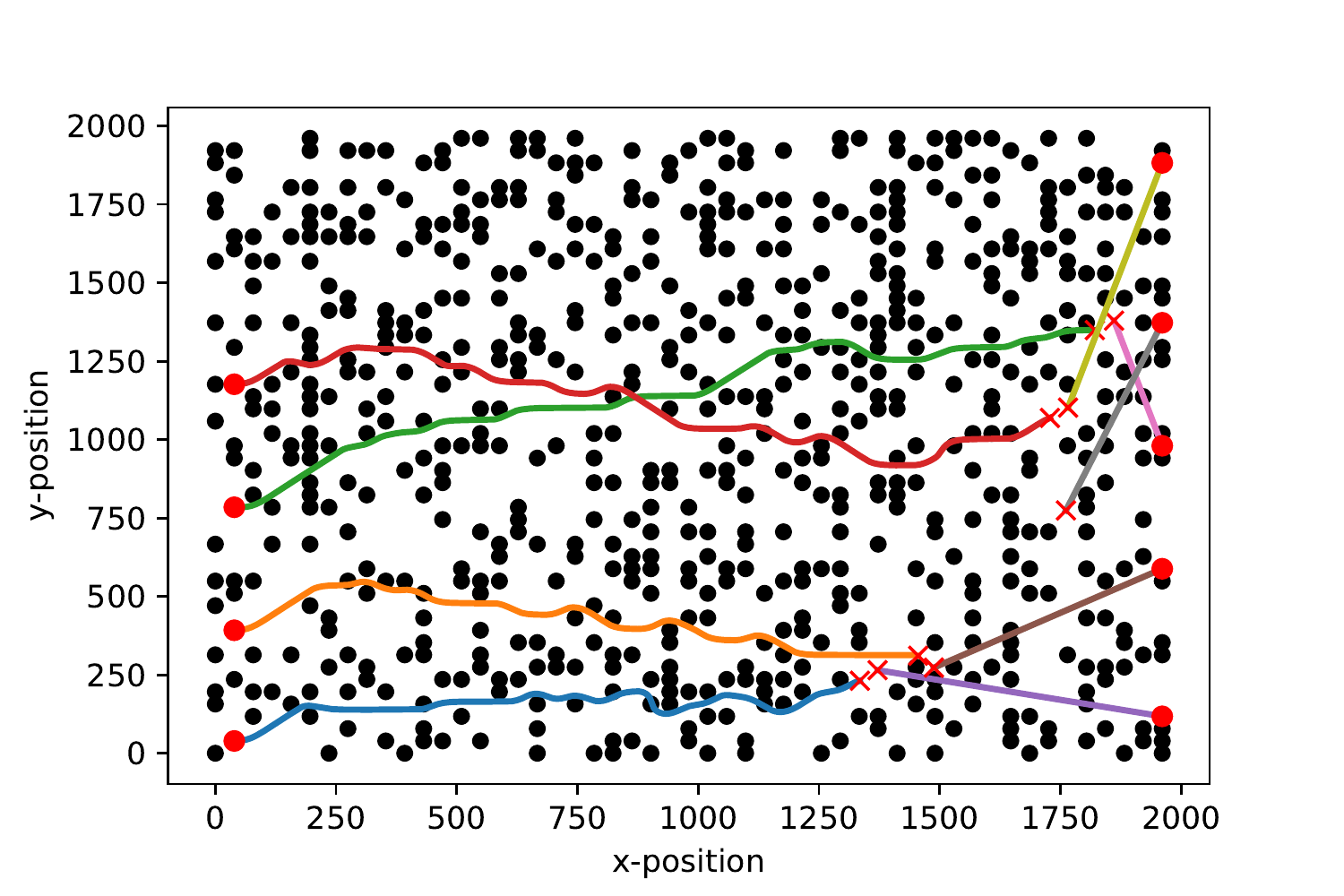}\\
	\caption{\textbf{Ground truth for diagonal target motion}}
\end{figure} 

\begin{figure}\label{OneColumn}
	\centering
	\includegraphics[width=3.25in]{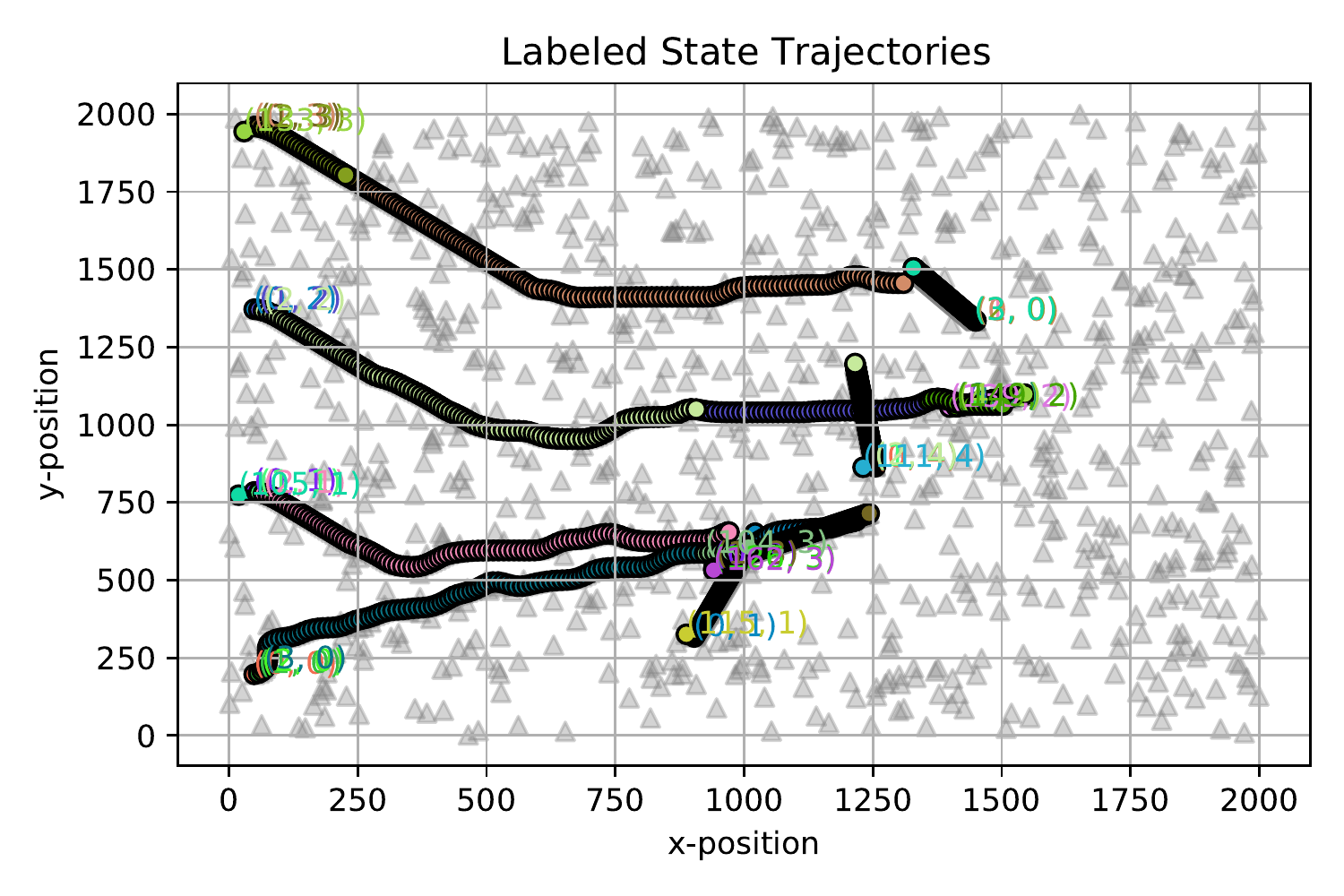}\\
	\caption{\textbf{GLMB output for random target location/motion}}
\end{figure}

\begin{figure}\label{OneColumn}
	\centering
	\includegraphics[width=3.25in]{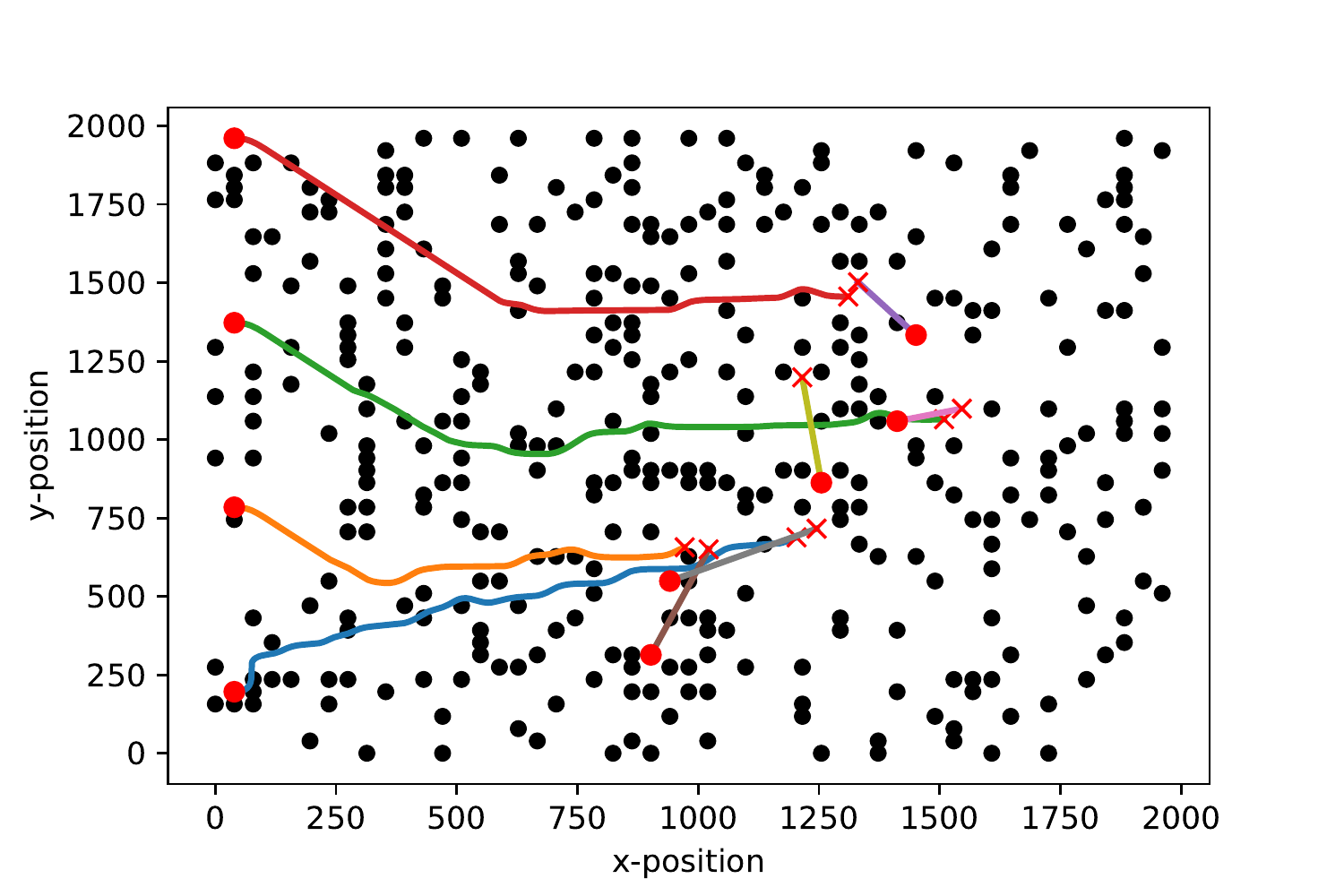}\\
	\caption{\textbf{Ground truth for random target location/motion}}
\end{figure}

\section{Conclusions}

A technique for autonomous UAS swarm dynamic mission planning was presented in this work. Simulations showed that the algorithm is robust to target motion, numerous obstacles in the mission area, and highly cluttered measurements.

This method can be used for arbitrarily many agents, but the computational requirements on consumer hardware become burdensome with more than five agents and targets due to the computational stress of the GLMB and $\mathcal{O}$$(n^{2})$ complexity in the A* loop. For slow-moving targets that do not move beyond the threshold to re-accomplish A* within one or two update periods this is not an issue, but fast-moving targets will cause A* to be invoked more frequently. This presents an issue for real-world applications where quick responses to fast-moving targets is crucial. Additionally there is quite a lot of room for runtime optimization, with potential for parallelization and more efficient language. However, the CPU used to generate these results was also burdened by 100 Hz dynamics updates and is of consumer grade. In applications central command could use a dedicated high-quality CPU that would only have to perform the GLMB filtering and A* swarm guidance updates to greatly increase execution speed.

Another issue with this implementation is the lack of target motion prediction. The current routine is not forward-looking, instead relying on target measurements to update the swarm guidance. The targets from this paper's results only move at approximately half the speed of the agents, but for targets moving at or above the agent velocity target state prediction is necessary because targets would be able to outrun agents. Results of such a simulation are shown in Figures 9 and 10. This problem is addressed in a parallel IEEE Aerospace 2021 paper by the second and third authors \cite{JDL_Thomas_IRL} that uses coupled Deep Q-learning and Inverse Reinforcement learning to estimate target dynamics purely from the GLMB output and predict target motion over a given time horizon. Combining that functionality with the techniques in this paper would improve performance when targets move slower than agents and allow mission completion with faster targets.

The results demonstrate good initial performance and promise for further development. Topics of interest to the authors include agent detection of new obstacles and targets not known to the human mission initiator, incorporating target importance and mission area danger to the A* calculations, target motion prediction, and different pathfinding algorithms such as Rapid Random Tree (RRT).

\begin{figure}\label{OneColumn}
	\centering
	\includegraphics[width=3.25in]{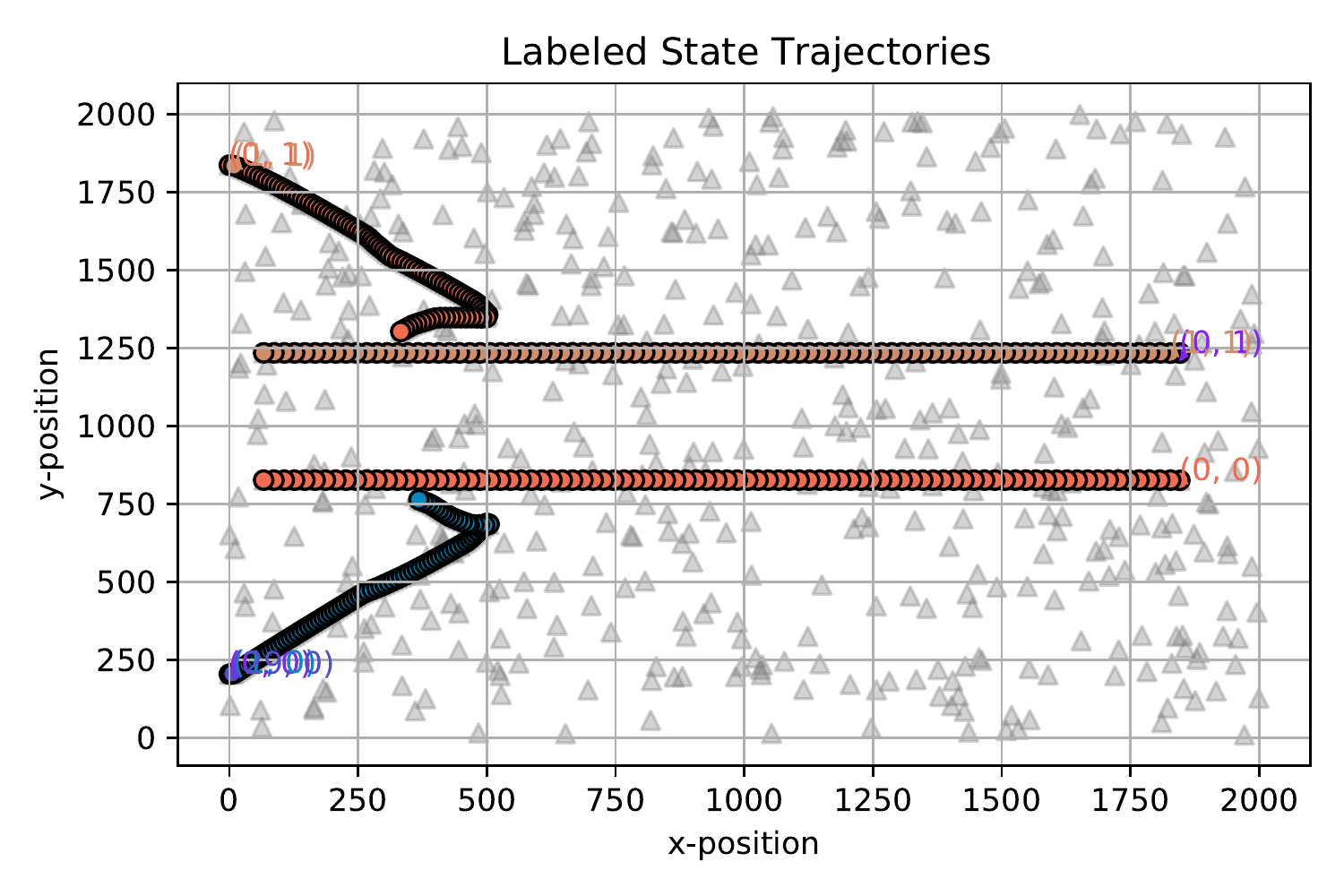}\\
	\caption{\textbf{GLMB output for fast target mission}}
\end{figure}

\begin{figure}\label{OneColumn}
	\centering
	\includegraphics[width=3.25in]{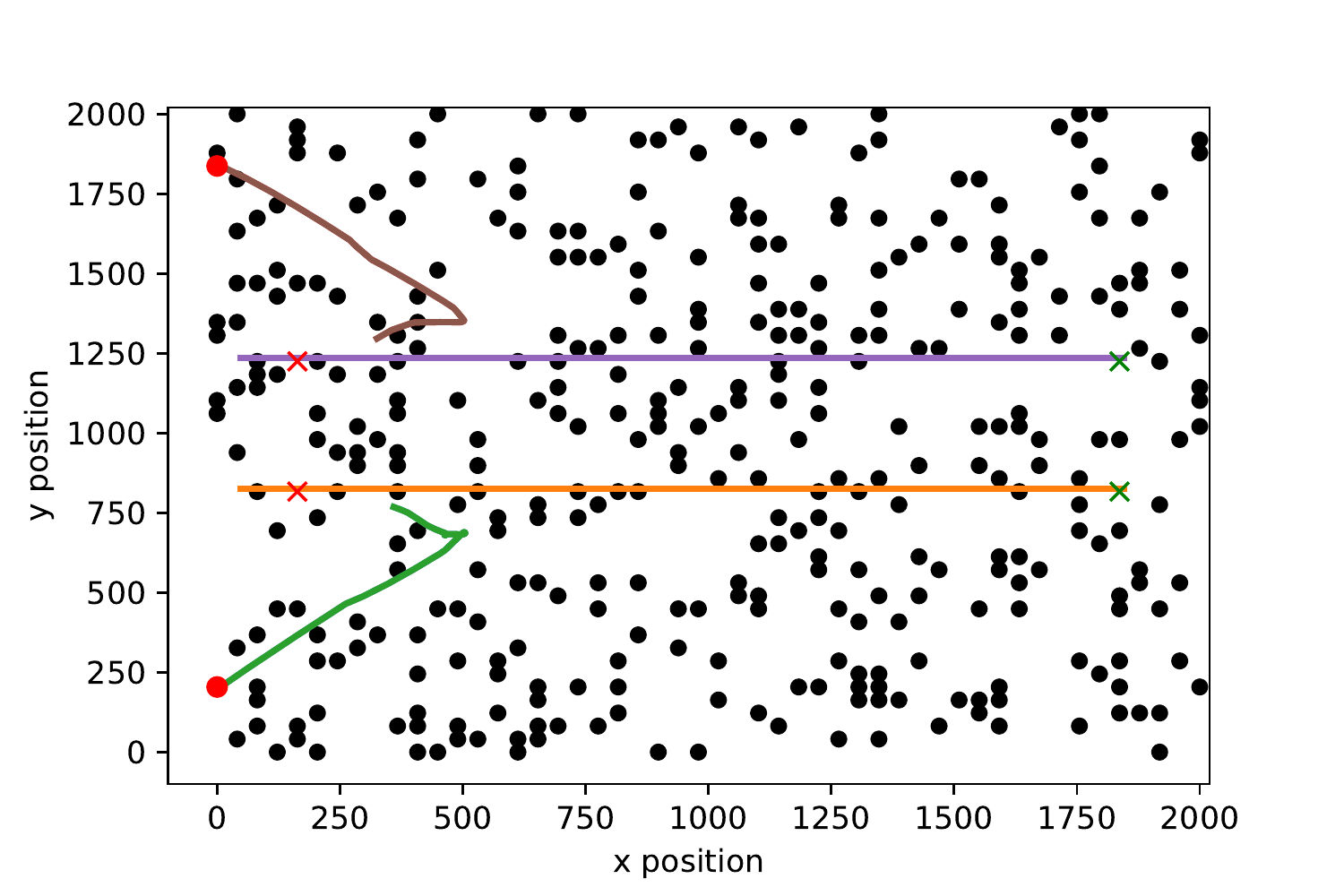}\\
	\caption{\textbf{Ground truth for fast target mission}}
\end{figure}

\bibliographystyle{IEEEtran}
\bibliography{\myreferences}

\begin{thebibliography}{10}
\providecommand{\url}[1]{#1}
\csname url@samestyle\endcsname
\providecommand{\newblock}{\relax}
\providecommand{\bibinfo}[2]{#2}
\providecommand{\BIBentrySTDinterwordspacing}{\spaceskip=0pt\relax}
\providecommand{\BIBentryALTinterwordstretchfactor}{4}
\providecommand{\BIBentryALTinterwordspacing}{\spaceskip=\fontdimen2\font plus
\BIBentryALTinterwordstretchfactor\fontdimen3\font minus
  \fontdimen4\font\relax}
\providecommand{\BIBforeignlanguage}[2]{{%
\expandafter\ifx\csname l@#1\endcsname\relax
\typeout{** WARNING: IEEEtran.bst: No hyphenation pattern has been}%
\typeout{** loaded for the language `#1'. Using the pattern for}%
\typeout{** the default language instead.}%
\else
\language=\csname l@#1\endcsname
\fi
#2}}
\providecommand{\BIBdecl}{\relax}
\BIBdecl

\bibitem{Vo2006_TheGaussianMixtureProbabilityHypothesisDensityFilter}
B.-N. Vo and W.-K. Ma, ``The gaussian mixture probability hypothesis density
  filter,'' \emph{IEEE Transactions on signal processing}, vol.~54, no.~11, pp.
  4091--4104, 2006.

\bibitem{Vo2006_TheCardinalizedProbabilityHypothesisDensityFilterforLinearGaussianMultiTargetModels}
B.-T. Vo, B.-N. Vo, and A.~Cantoni, ``The cardinalized probability hypothesis
  density filter for linear gaussian multi-target models,'' in \emph{2006 40th
  Annual Conference on Information Sciences and Systems}.\hskip 1em plus 0.5em
  minus 0.4em\relax IEEE, 2006, pp. 681--686.

\bibitem{Vo2013_LabeledRandomFiniteSetsandMultiObjectConjugatePriors}
B.-T. Vo and B.-N. Vo, ``Labeled random finite sets and multi-object conjugate
  priors,'' \emph{IEEE Transactions on Signal Processing}, vol.~61, no.~13, pp.
  3460--3475, Jul 2013.

\bibitem{Reuter2014_TheLabeledMultiBernoulliFilter}
S.~Reuter, B.-T. Vo, B.-N. Vo, and K.~Dietmayer, ``The labeled multi-bernoulli
  filter,'' \emph{IEEE Transactions on Signal Processing}, vol.~62, no.~12, pp.
  3246--3260, 2014.

\bibitem{Vo2014_LabeledRandomFiniteSetsandtheBayesMultiTargetTrackingFilter}
B.-T. Vo, B.-N. Vo, and D.~Phung, ``Labeled random finite sets and the bayes
  multi-target tracking filter,'' \emph{IEEE Transactions on Signal
  Processing}, vol.~62, no.~24, pp. 6554--6567, Dec 2014.

\bibitem{gasur}
\BIBentryALTinterwordspacing
J.~D. Larson, R.~W. Thomas, and V.~Weirens, ``Gasur: a python library for
  guidance, navigation, and control of autonomous swarms using random finite
  sets,'' Web page, 2019. [Online]. Available:
  \url{https://github.com/drjdlarson/gasur}
\BIBentrySTDinterwordspacing

\bibitem{gncpy}
\BIBentryALTinterwordspacing
J.~D. Larson and R.~W. Thomas, ``Gncpy: A python library for guidance,
  navigation, and control algorithms,'' Web page, 2019. [Online]. Available:
  \url{https://github.com/drjdlarson/gncpy}
\BIBentrySTDinterwordspacing

\end{thebibliography}


\begin{thebibliography}{10}
\providecommand{\url}[1]{#1}
\csname url@samestyle\endcsname
\providecommand{\newblock}{\relax}
\providecommand{\bibinfo}[2]{#2}
\providecommand{\BIBentrySTDinterwordspacing}{\spaceskip=0pt\relax}
\providecommand{\BIBentryALTinterwordstretchfactor}{4}
\providecommand{\BIBentryALTinterwordspacing}{\spaceskip=\fontdimen2\font plus
\BIBentryALTinterwordstretchfactor\fontdimen3\font minus
  \fontdimen4\font\relax}
\providecommand{\BIBforeignlanguage}[2]{{%
\expandafter\ifx\csname l@#1\endcsname\relax
\typeout{** WARNING: IEEEtran.bst: No hyphenation pattern has been}%
\typeout{** loaded for the language `#1'. Using the pattern for}%
\typeout{** the default language instead.}%
\else
\language=\csname l@#1\endcsname
\fi
#2}}
\providecommand{\BIBdecl}{\relax}
\BIBdecl

\bibitem{Mahler2003_MultiTargetBayesFiltering}
R.~Mahler, ``Multi-target Bayes filtering via first-order
multi-target moments,'' \emph{IEEE Transactions on AES}, 
vol.~39, no.~4, pp. 1152--1178, 2003.

\bibitem{Vo2006_TheGaussianMixtureProbabilityHypothesisDensityFilter}
B.-N. Vo and W.-K. Ma, ``The gaussian mixture probability hypothesis density
  filter,'' \emph{IEEE Transactions on signal processing}, vol.~54, no.~11, pp.
  4091--4104, 2006.

\bibitem{Vo2006_TheCardinalizedProbabilityHypothesisDensityFilterforLinearGaussianMultiTargetModels}
B.-T. Vo, B.-N. Vo, and A.~Cantoni, ``The cardinalized probability hypothesis
  density filter for linear gaussian multi-target models,'' in \emph{2006 40th
  Annual Conference on Information Sciences and Systems}.\hskip 1em plus 0.5em
  minus 0.4em\relax IEEE, 2006, pp. 681--686.

\bibitem{Vo2013_LabeledRandomFiniteSetsandMultiObjectConjugatePriors}
B.-T. Vo and B.-N. Vo, ``Labeled random finite sets and multi-object conjugate
  priors,'' \emph{IEEE Transactions on Signal Processing}, vol.~61, no.~13, pp.
  3460--3475, Jul 2013.

\bibitem{Reuter2014_TheLabeledMultiBernoulliFilter}
S.~Reuter, B.-T. Vo, B.-N. Vo, and K.~Dietmayer, ``The labeled multi-bernoulli
  filter,'' \emph{IEEE Transactions on Signal Processing}, vol.~62, no.~12, pp.
  3246--3260, 2014.

\bibitem{Vo2014_LabeledRandomFiniteSetsandtheBayesMultiTargetTrackingFilter}
B.-T. Vo, B.-N. Vo, and D.~Phung, ``Labeled random finite sets and the bayes
  multi-target tracking filter,'' \emph{IEEE Transactions on Signal
  Processing}, vol.~62, no.~24, pp. 6554--6567, Dec 2014.

\bibitem{astar}
P.E.~Hart, N.J.~Nilsson, and B.~Raphael, ``A Formal Basis for the Heuristic Determination 
of Minimum Cost Paths,'' \emph{IEEE Transactions on Systems Science and Cybernetics,}, 
vol.~39, pp. 100--107, 1968.

\bibitem{nilsson}
N.J.~Nilsson, ``The Quest for Artificial Intelligence,'' \emph{Cambridge University Press,}, 
2010.

\bibitem{Kuhn_hungarian}
H.W.~Kuhn, ``The Hungarian Method for the Assignment Problem,'' \emph{Naval Research Logistics Quarterly,} 
pp 83-87, 1955.

\bibitem{munkres}
J.~Munkres, ``Algorithms for the Assignment and Transporation Problems,'' \emph{Journal of the Society for 
Industrial and Applied Mathematics,} pp 32-38, 1957.

\bibitem{JDL_Linares_ION}
J.D.~Larson, B.~Doerr, and R.~Linares, ``Autonomous
Mission Planning for Swarms Using Random Finite
Sets,'' \emph{Proceedings of the 32nd International Technical
Meeting of the Satellite Division of The Institute of
Navigation (ION GNSS+ 2019)} pp 1753-1761, 2019.

\bibitem{JDL_Thomas_ELQR}
R.W.~Thomas and J.D.~Larson, ``Receding Horizon
Extended Linear Quadratic Regulator for RFS-based
Swarms with Target Planning and Automatic Cost Function Scaling'' \emph{IEEE,}
to be published.

\bibitem{JDL_Thomas_IRL}
R.W.~Thomas and J.D.~Larson, ``Inverse Reinforcement Learning for Generalized Labeled Multi-Bernoulli
Multi-Target Tracking,'' \emph{IEEE Aerospace 2021,}
to be published.

\bibitem{gasur}
\BIBentryALTinterwordspacing
J.~D. Larson, R.~W. Thomas, and V.~Weirens, ``{GASUR}: A {P}ython library for
  {G}uidance, navigation, and control of {A}utonomous {S}warms {U}sing {R}andom
  finite sets,'' Web page, 2019. [Online]. Available:
  \url{https://github.com/drjdlarson/gasur}
\BIBentrySTDinterwordspacing

\bibitem{gncpy}
\BIBentryALTinterwordspacing
J.~D. Larson and R.~W. Thomas, ``{GNCPy}: A {P}ython library for {G}uidance,
  {N}avigation, and {C}ontrol algorithms,'' Web page, 2019. [Online].
  Available: \url{https://github.com/drjdlarson/gncpy}
\BIBentrySTDinterwordspacing

\end{thebibliography}

\thebiography

\begin{biographywithpic}
	{Vincent W. Hill}{hillvi}
	received a B.S. in mechanical engineering and a M.S. degree in aerospace engineering and mechanics from the University of Alabama, Tuscaloosa, AL, USA in 2017 and 2020, respectively. His research interests include autonomous systems, multi-agent/multi-sensor systems, and aeroservoelasticity. In parallel with pursuing his Ph.D he is currently employed with Aerovironment, inc. in Simi Valley, CA as a GNC Engineer on the solar high-altitude long-endurance aircraft program.
	
\end{biographywithpic}

\begin{biographywithpic}
	{Jordan D. Larson}{larso}
	received a B.S., M.S., and Ph.D. degrees in aerospace engineering from University of Minnesota, Minneapolis, MN, USA in 2013, 2015, and 2018, respectively. From 2018 to 2019, he was a Post Doctoral Research Associate with the University of Minnesota before becoming an assistant professor at the University of Alabama. His research interests include autonomous systems, guidance, navigation, and control, multi-agent/multi-sensor systems, signal processing, and uncertainty quantification. Dr. Larson also recently received the Samuel M. Burka Award from the Institute of Navigation.
	
\end{biographywithpic}

\begin{biographywithpic}
	{Ryan W. Thomas}{thoma}
	received a B.S. degree in aerospace engineering from the University of Minnesota, Minneapolis, MN, USA in 2019. His research interests include autonomous systems, guidance, navigation, and control, and multi-agent/multi-sensor systems.
	
\end{biographywithpic}

\end{document}